# Vehicle Safety Management System

**Chanthini Bhaskar[1], Bharath Manoj Nair[2], Dev Mehta[3]**
[1]School of Electronics Engineering, Vellore Institute of Technology, Vandalur, Chennai, Tamil Nadu, India - 600127

Corresponding author: Chanthini Bhaskar (e-mail: chanthini.baskar@vit.ac.in).

**ABSTRACT** Overtaking is a critical maneuver in driving that requires accurate information about the location and distance of other vehicles on the road. This study suggests a real-time overtaking assistance system that uses a combination of the You Only Look Once (YOLO) object detection algorithm and stereo vision techniques to accurately identify and locate vehicles in front of the driver, and estimate their distance. The system then signals the vehicles behind the driver using colored lights to inform them of the safe overtaking distance. The proposed system has been implemented using Stereo vision for distance analysis and You Only Look Once (YOLO) for object identification. The results demonstrate its effectiveness in providing vehicle type and the distance between the camera module and the vehicle accurately with an approximate error of 4.107%. Our system has the potential to reduce the risk of accidents and improve the safety of overtaking maneuvers, especially on busy highways and roads.

## I. INTRODUCTION

The issue of road safety has been of critical importance globally, as road traffic accidents have caused significant harm to individuals and society. Road traffic accidents are estimated to be the cause of 1.35 million deaths annually, according to the World Health Organization[1], making them the eighth leading cause of death worldwide. Overtaking has been one of the key contributors to road accidents, particularly on highways and rural roads. To address this problem, a new overtaking assistance system was proposed in this research paper that provided clear and timely signals to vehicles behind the host vehicle.

In recent years, advances in computer vision and sensing technologies have led to the development of intelligent transportation systems that could enhance driving safety and efficiency. In this paper, a real-time overtaking assistance system was proposed that used a combination of the You Only Look Once (YOLO) object detection algorithm and stereo vision techniques to accurately identify and locate vehicles in front of the driver and estimate their distance. Using this information, the vehicles behind the host vehicle were signaled by means of colored signals, helping drivers make informed decisions on the overtaking procedure.

The system was developed and tested in real-world scenarios, and the results demonstrated its effectiveness in detecting vehicles and accurately calculating their distances. The details of the system design and testing will be presented in this research paper, and the potential benefits and limitations of the approach will also be discussed.

## II. LITERATURE REVIEW

In recent years, there has been a growing interest in understanding the issues that are causes of accidents on the road and developing solutions for them. One of the developed solutions include the Advanced Driver Assistance Systems (ADAS) to reduce the risk of accidents on the road. There have also been some attempts to understand driver behavior in overtaking accidents and factors, both human and external, that play into the probability of such accidents.

Ravindra Kumar Kinikeri et al. studied the causes of vehicle crashes due to human error. They discovered that out of 206 accidents, 64% were classified as either fatal or serious, and 170 of them were recorded with clear weather conditions. 70 of these were due to no driver distraction, 44 were due to sleepy conditions, and 45 were due to other reasons. 62 of these accidents were filtered for clear weather, no junction, no driver distraction, and no vision obstruction. Analysis of the contributing factors in these 62 accidents concluded that 51% of the accidents were due to human error, 36% due to vehicle error, and 13% due to infrastructure error. [2]

Hasmar Halim et al. presented a model for predicting the risk of accidents during motorcycle overtakingIt was discovered that the frequency of accidents dropped as vehicle distance increased. However, as the vehicle distance increased, the chance of an accident increased due to several factors in the lateral distance. Accordingly, the space between two moving objects was recommended to be 1.88 meters in the latitudinal direction and 4.62 meters in the longitudinal direction. [3]

Jun-Ming Xu et al. proposed a mechanism that combined a VTT module along with a distance measurement technique to signal rear vehicles and help them make informed decisions. [4]

Sheng-Xiu Lin proposed an in-vehicle safety-aided system that used DC power-line communication to provide real-time safety alerts to drivers. They highlighted the potential of this system to enhance road safety by providing drivers with accurate and timely information about road conditions. [5]

Schittenhelm Helmut put forth a driver assistance system that could be used in oncoming traffic situations to reduce the risk of accidents. Dr. Helmut continued by saying that warnings roughly lowered the crash rate by 70%. It was discovered that when simple cautions were provided, drivers were in a much better position to judge and evaluate the circumstances. Thus, there were significantly fewer frantic reactions, crashes, or near misses. [6]

Linghe Kong focused on millimeter-wave wireless communications and their potential to support autonomous vehicles. They proposed a solution for an IoT-cloud backed vehicular mmWave system that fully utilised mmWave and vehicle benefits. On one hand, sensory data could be shared by cars using this system to combat blind spots and weather-related issues. On the other hand, human movements and small objects could be accurately recognised using cloud computing via V2I transmission of HD video. [7]

Ming Li et al. made an effort to examine the trajectory of lane changes made by vehicles and offered a model for the crucial instant when one vehicle passes another. According to the results of the CarSim simulation, the crucial time point was shrunk as lateral distance and longitudinal relative speed between the two vehicles increased. The system architecture greatly aided the development of active safety measures for future driverless cars. [8]

"Research on Modeling and Optimization of Overtaking Rules under Intelligent System" proposed a model for optimizing overtaking rules. Here, Hai Liu et al. presented a model that could be used to optimize overtaking rules based on vehicle speed, distance, and other factors, thereby reducing the risk of accidents during overtaking. [9]

Alexey Vinel in "An Overtaking Assistance System Based on Joint Beaconing and Real-Time Video Transmission" proposed an overtaking assistance system for vehicles. The system utilized a combination of two technologies: joint beaconing and real-time video transmission. Joint beaconing was used to establish a communication link between the two vehicles. When the driver of the following vehicle decided to overtake the leading vehicle, the system activated the camera mounted on the back of the leading vehicle. The video footage from the camera was then transmitted in real-time to the driver of the following vehicle, providing them with a clear view of the road ahead. The system also included a warning mechanism that alerted the driver of the following vehicle if it was not safe to overtake. The warning was based on data obtained from the joint beaconing protocol, such as the distance between the two vehicles and their relative speeds. [10]

An important paper that was looked into was "Opportunities of advanced driver assistance systems towards overtaking" by Geertje Hegeman. This paper had done in-depth research on the overtaking maneuver done by several drivers and had tried to narrow down the entire process into different subtasks, through which difficult and dangerous subtasks could be analyzed. They also analyzed the average time taken for the overtake maneuver and the time taken for overtake decision.[11]

The potential benefits of utilizing H.264/AVC, IEEE 802.11p/WAVE, and automotive radars in video-based overtaking assistance applications were demonstrated in "The Use of Automotive Radars in Video-Based Overtaking Assistance Applications" by Evgeny Belyaev. The paper discussed the application of automotive radars and their potential in combination with other technologies. [12]

### III. METHODOLOGY

1) HARDWARE PARTS ACQUIRED

A Raspberry Pi 4, a compact and affordable single-board computer that is frequently used for a variety of embedded projects, is the platform on which the suggested overtaking aid system is intended to be implemented. Two HP web cameras are employed in the system as input hardware. The Raspberry Pi 4 and the cameras, which are fixed to the front of the vehicle, are linked by USB Type-A connections. The stereo vision algorithm uses the stereo images that the cameras have taken of the road ahead to determine how far the vehicles in front of the driver are from it. The project can be extended to include the use of an array of 5 LEDs for the purpose of signaling. For the purpose of this project as a proof of concept, we shall display the results using a monitor.

We will go into further details regarding the hardware components in the sections below.

- Raspberry PI 4: Raspberry Pi 4 is a versatile computer for various projects, with quad-core processor, up to 8GB RAM, GPIO pins, and flexible OS support.
- HP w100 480P 30 FPS Digital Webcam: HP w100 webcam is affordable, compact and captures video at 480p/30FPS. It's suitable for personal/professional use, and used in project.
- Micro HDMI: Micro HDMI is a smaller version of HDMI that transmits audio and video between devices. Raspberry Pi 4 has 2 micro HDMI ports.
- USB Type C: USB Type-C is a versatile and fast USB connector that delivers up to 100 watts. Raspberry Pi has a Type-C charging port.

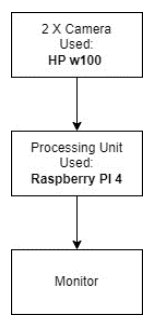

FIGURE 01.     Hardware block diagram of the system

2) SOFTWARE PARTS ACQUIRED

For our project, we have chosen Python and its libraries. The project is developed on Raspberry PI OS and implemented using Visual Code Studio. Each of these software tools serves a specific purpose and helps in achieving the desired outcome of the project.

- Raspberry PI OS: Raspberry Pi OS is a free, lightweight operating system designed for the Raspberry Pi computer.
- Visual Code Studio: A free source-code editor developed by Microsoft for different operating systems.
- Python and its libraries: Python is popular for IoT due to its wide range of libraries and tutorials, including essential ones like NumPy and Pandas.

3) BASIC OVERVIEW

Overtaking is a critical maneuver in driving that requires accurate information about the location and distance of other vehicles on the road. A driver must consider several factors before making an overtake, including the speed of their vehicle, the speed of the vehicle being overtaken, the distance to oncoming traffic, and the visibility of the road ahead. In this project, we aim to solve the visibility and the distance aspect of the problem. Most drivers only have a time frame of less than 1 second for overtake decision and around 7.8 seconds to complete the maneuver. We aim to create a system as mentioned in the diagram given below.

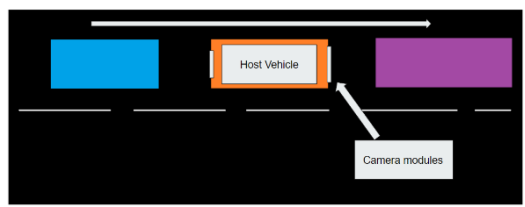

**FIGURE 02.**     **Basic overview**

As seen in figure 02, we aim to create a system that implements the use of two forward facing camera, which will be fitted in the front of the host vehicle, and a LED array can be used to help signal the vehicle behind it.

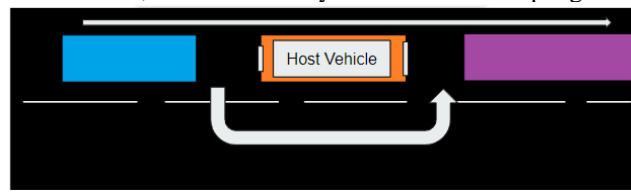

**FIGURE 03.**     **Overtake procedure**

Referring to figure 03, our goal is to implement a system that will help the vehicle(blue) which is behind the host vehicle overtake the latter safely. To ensure this, there are several factors that must be present within the model. These factors include:

- Identify type of vehicle that is in front of host vehicle ( Orange )
- Identify distance between host vehicle (orange) and the vehicle(Purple) in front of it..

The front facing camera module is responsible for capturing the vehicle in front of it, in two different perspectives, from which is then passed through You Only Look Once (YOLO) algorithm and Stereo-Vision algorithm for object detection and distance calculation. This then can be signaled to the vehicle (Blue) behind the host vehicle via the LED array or any other signaling.

4) SOFTWARE OVERVIEW

The two main pillars of the software aspect of the project are the You Only Look Once (YOLO) algorithm and the stereo-vision algorithm. An in-depth description of the algorithm is as given below.

a) You Only Look Once (YOLO) : You Only Look Once (YOLO) is a deep learning technique for object detection, which was developed by Joseph Redmon, Ali Farhadi, and others in 2016. You Only Look Once (YOLO) is a real-time object detection system that can detect objects in images and videos with high accuracy and speed. Based on a single convolutional neural network, the You Only Look Once (YOLO) technique can forecast bounding boxes and class probabilities for objects in an input picture or video frame.. A convolutional neural network is applied to each cell in the grid-like structure that the You Only Look Once (YOLO) algorithm creates from a divided input image. Each cell in the grid predicts a set of bounding boxes, which represent potential objects in the image. Each bounding box consists of four values: the x coordinate and y coordinate of the center of the box, the width of the box, and the height of the box. The neural network also predicts a probability for each bounding box, indicating the likelihood that the box contains an object of a particular class. You Only Look Once (YOLO) is excellent for applications that require real-time object detection, such as autonomous vehicles, surveillance systems, and robotics. You Only Look Once (YOLO) has been used in this project for object location and detection.

b) Stereo-Vision: Stereo vision is a technique used in computer vision to estimate the 3D distance of objects from a pair of 2D images taken from slightly different viewpoints. Stereo vision is based on the principle of triangulation, which involves using the displacement of corresponding points in two images to calculate the depth of objects in the scene. In stereo vision, two cameras are placed side-by-side and capture images of the same scene from slightly different viewpoints. The distance between the two cameras is called the baseline. The images captured by the cameras are then processed to extract corresponding points or features, which are points in the two images that represent the same physical point in the scene. Once the corresponding points are identified, the next step is to calculate the disparity, which is the horizontal distance between the corresponding points in the two images. The disparity is directly proportional to the depth of the objects in the scene, and can be calculated using the following formula:

$$disparity = xL - xR \qquad (1)$$

where xL and xR are the horizontal coordinates of the corresponding points in the left and right images, respectively.

Once the disparity is calculated, the depth of the objects in the scene can be estimated using the triangulation principle. The depth of an object can be calculated using the following formula:

$$depth = (baseline * f\ length)/disparity \qquad (2)$$

where baseline is the distance between the two cameras, and f length or focal length is the distance between the camera lens and the image sensor.

One of the advantages of stereo vision is that it is a passive sensing technique that does not require any additional illumination or active sensors. This makes stereo vision suitable for applications such as autonomous vehicles, robotics, and surveillance systems, where active sensors such as LiDAR or radar may not be feasible.

5) SOFTWARE FRAMEWORK

As explained in the section above, the software implementation is a combination of You Only Look Once (YOLO) and stereo-vision algorithm carried out in the Python language. Object detection and localization of the image obtained from both cameras is done by the You Only Look Once (YOLO) algorithm, the localized coordinates are then used to calculate distance. A general flowchart is as given below.

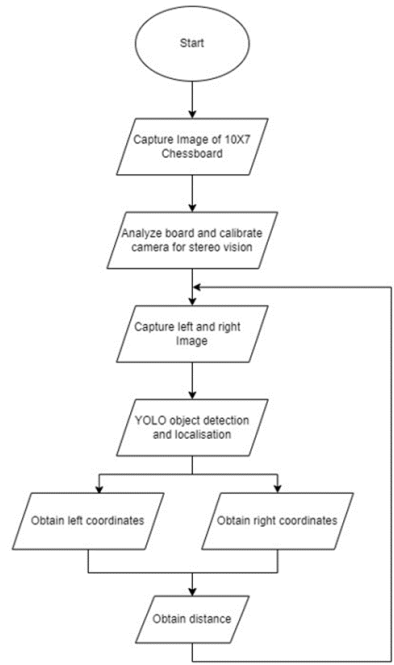

**FIGURE 04.** User interaction Software flow chart

The user is only responsible for basic calibration of the model. Within the model however, there are 5 sub-programs that are responsible for the software implementation.

a) calibration_image.py: The code imports the OpenCV library and creates two instances of the VideoCapture class to capture video frames from two cameras, one from camera 1 and another from camera 0. A while loop is initiated, which continues to execute as long as the capture from camera 1 is active. Within the loop, the code reads a frame from both cameras and displays them in separate windows. It also waits for a key press for 5 milliseconds, and if the pressed key is 's', the current frames from both cameras are saved as images in a specific format within a designated folder. If the pressed key is 'ESC', the while loop breaks and the program exits.

b) stereo_calibration.py: The code performs stereo camera calibration to obtain camera matrices, distortion coefficients, rotation and translation vectors, and other parameters required for rectifying the stereo images and calculating depth information. It imports necessary libraries, including numpy, cv2, and glob. The code sets the size of the chessboard and frame used for calibration. The criteria for sub-pixel accuracy of corner detection is defined. The code generates 3D object points and stores them. Left and right stereo images are loaded, and chessboard corners are detected in each image. The object points and corresponding image points of the corners are saved. Stereo calibration is performed to obtain the necessary parameters. The stereo images are rectified to remove distortion and align them for stereo correspondence. A stereo map is created for each rectified image to map pixels from one camera to the other. The stereo map data is saved in a file named "stereoCalibration.xml".

c) calibration.py: The code imports necessary libraries, including sys, numpy, time, imutils, and OpenCV. It opens the 'stereoCalibration.xml' file using the cv2.FileStorage class. The code reads the stereoMapL_x, stereoMapL_y, stereoMapR_x, and stereoMapR_y matrices from the file using the getNode() method. 'stereoMapL_x, stereoMapL_y, stereoMapR_x, and stereoMapR_y are variables that contain maps used to undistort and rectify images captured by a stereo camera setup. Undistortion and rectification are essential steps in stereo vision, where the goal is to extract depth information from two images captured by cameras placed at different positions. The maps contain information on how each pixel in the original image should be transformed to compensate for any distortion and alignment issues caused by the camera lenses and their positions. These maps are generated during the stereo calibration process, where the positions and orientations of the cameras are determined, and then used to undistort and rectify images in subsequent processing steps. A function named undistortRectify is defined that takes two frames, frameR and frameL, as arguments. Inside the function, the frames are undistorted and rectified using the cv2.remap() function with the stereoMap matrices. Finally, the function returns the undistorted and rectified frames, undistortedR and undistortedL, respectively.

d) stereo_vision.py: The code imports required modules such as cv2, numpy, time, and matplotlib. It defines two instances of the VideoCapture class to capture video frames from the two cameras. The frame rate of both cameras is set to 20 frames per second. The code initializes the values of the variables B, f, and alpha. The images obtained are also undistorted based off the parameters received from the calibration configuration program. The You Only Look Once (YOLO) library is used to detect objects in the frames captured from both cameras. The frames are calibrated to undistort and rectify them. Bounding boxes are drawn around the detected objects. The depth of the object is calculated using the coordinates obtained and the parameters B, f, and alpha which are fed into the triangulation section.
e) triangulation.py: The code defines a function find_depth() that takes in the following inputs:

- right_point: The x and y pixel coordinates of the point in the right frame
- left_point: The x and y pixel coordinates of the point in the left frame
- frame_right: The right frame of the stereo camera
- frame_left: The left frame of the stereo camera
- baseline: The baseline distance between the cameras
- f: Focal length of the cameras
- alpha: The angle of view of camera

The function converts focal length f from [mm] to [pixel], which is stored as f_pixel, computes the pixel disparity (stored as disparity) between the two frames, and then calculates the depth in [cm] using the formula:

$$depth = (baseline * f\_pixel)/disparity$$

The function returns the absolute value of the depth in [cm].

The flowchart given below further explains flow of control through the sub-programs within the software implementation.

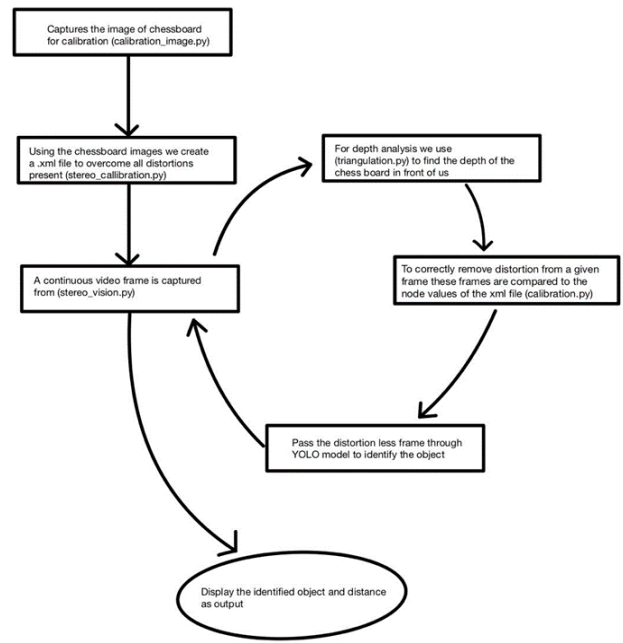

FIGURE 05.   Software flowchart

## IV. RESULTS AND DISCUSSION

### A. RESULTS

The Stereo-Vision Technique using You Only Look Once (YOLO) algorithm for localization model as was tested with several types of vehicles, The model not only classified the vehicles but also gave a fair estimated distance to the object. The model was, however, seen to have an average latency of about 0.4755 seconds. This reduced the future scope of its use in acceleration detection and the speed of recognition. A test of the model gives the following values.

TABLE I
ERROR OF DISTANCE MEASURED

| Distance measured | | | |
|---|---|---|---|
| Actual (cm) | Measured (cm) | Difference (cm) | Error (as a percent) |
| 34 | 36 | 2 | 5.88 |
| 35 | 37 | 2 | 5.71 |
| 40 | 40 | 0 | 0 |
| 42 | 45 | 3 | 7.14 |
| 43 | 45 | 2 | 4.65 |
| 45 | 48 | 3 | 6.66 |
| 52 | 51 | 1 | 1.92 |
| 54 | 56 | 2 | 3.70 |
| 55 | 53 | 2 | 3.63 |
| 56 | 55 | 1 | 1.78 |

As we can see, the model has an approximate error percentage of 4.107%. The model was also able to accurately depict the type of vehicle present.

Since it was not viable to move the model, we have used pictures of different types of vehicles to check for object detection. The model accurately depicted cars and bikes.

Far distance object recognition also posed a problem, as the cameras used in the project were not self-focusing, which often led to the vehicles not being captured due to obscured image caught.

### B. HARDWARE SETUP IMAGES

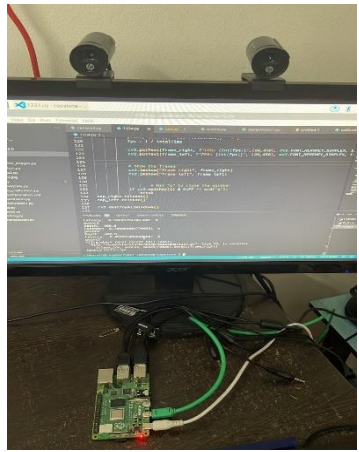

FIGURE 06.    Image of hardware setup

### C. Screenshots from console

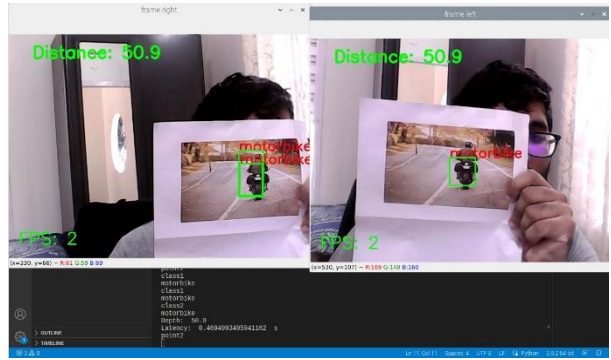

**FIGURE 07.** Monitor output for motorbike

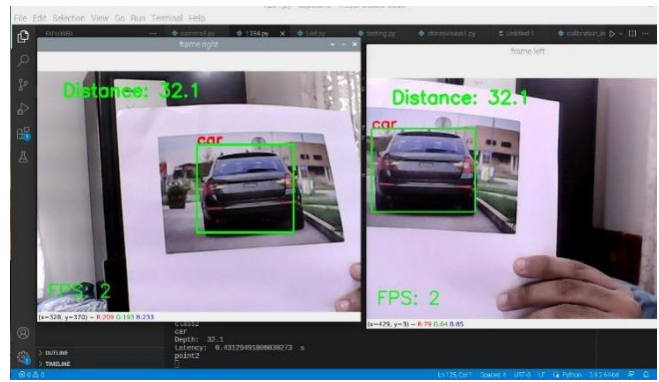

**FIGURE 08.** Monitor output for car

### D. Discussion

During the testing phase, several advantages and disadvantages have been recorded in the model. The first model (Stereo-Vision Technique using You Only Look Once (YOLO) algorithm for localization) brought about a robust method of detection, however, it also brought about the issue of high latency. This is due to the continuous use of the You Only Look Once (YOLO) algorithm for localization. One possible way, we can overcome this issue would be to use multiple ultrasonic for distance calculation, and relegated the Raspberry PI model to only object recognition. With the improved latency with regards to distance calculation, we can also indicate the state of acceleration of the vehicle in front of the host vehicle, which is an important point of consideration for the overtaking vehicle, which is behind the host vehicle.

To mitigate the issue faced in the method, we approached the problem via a second method (Stereo-Vision Technique using number plate recognition for localization), we saw that the model picked up and tracked any object that resembled a square or a rectangle, which might not always be a number plate. This can be very concerning if we implement wide-angle cameras for the project, which will also pick up rectangle-shaped objects like road signs, highway dividers, etc. One possible way we can solve this issue is to implement an object detection model and train it to detect the number plate from the contour locations provided by the model. But this once again brings us back to the issue faced in the first model, which was high latency. Thus, keeping in mind the factors brought about by both the models, we decided to use the You Only Look Once (YOLO) algorithm as the localization technique

### V. CONCLUSION

The research paper titled "Vehicle Safety Management System" proposed a real-time overtaking assistance system that used a combination of You Only Look Once (YOLO) object detection algorithm and stereo vision techniques to accurately identify and locate vehicles in front of the driver, and estimate their distance. Here, Raspberry PI 4 model B was used for processing, and two external cameras were used as input hardware. Two main algorithms were used here, You Only Look Once (YOLO) and Stereo-Vision, as the main foundation of the project. The results showed that we were able to accurately detect the type of vehicle and the distance to the vehicle within an error of about 4.107%. The system also showed a latency of up to 0.4755 seconds. We have also discussed the various advantages and disadvantages of the model as well as the reasons why we have selected the algorithms.

We have looked into an alternate technique to solve inherent issues present with our model and have stated issues and how to tackle them.